\documentclass[aps,pra,twocolumn,tightenlines,floatfix,showpacs,superscriptaddress,10pt]{revtex4-1}

\usepackage{amsmath}
\usepackage{amsfonts}
\usepackage{amssymb}
\usepackage{graphicx}%
\usepackage{xcolor}
\usepackage[percent]{overpic}
\providecommand{\U}[1]{\protect\rule{.1in}{.1in}}
\begin{document}
	
\author{Fabio L. Traversa}
\email{email: ftraversa@memcpu.com}
%\affiliation{Department of Physics, University of California, San Diego, La Jolla, 92093 CA}
\affiliation{MemComputing, Inc., San Diego, CA, 92130 CA}

\author{Pietro Cicotti}
\email{email: pcicotti@sdsc.edu}
\affiliation{San Diego Supercomputer Center, La Jolla, 92093 CA}

\author{Forrest Sheldon}
\affiliation{Department of Physics, University of California, San Diego, La Jolla, 92093 CA}

\author{Massimiliano Di Ventra}
\email{email: diventra@physics.ucsd.edu}
\affiliation{Department of Physics, University of California, San Diego, La Jolla, 92093 CA}

\title{Evidence of an exponential speed-up in the solution of hard optimization problems}

\maketitle

\textbf{
Optimization problems pervade essentially every scientific discipline and industry. Many such problems require finding a solution that maximizes the number of constraints satisfied. Often, these problems are particularly difficult to solve because they belong to the NP-hard class, namely algorithms that always find a solution in polynomial time are not known. Over the past decades, research has focused on developing heuristic approaches that attempt to find an approximation to the solution. However, despite numerous research efforts, in many cases even approximations to the optimal solution are hard to find, as the computational time for further refining a candidate solution grows exponentially with input size. Here, we show a {\it non-combinatorial} approach to hard optimization problems that achieves an {\it exponential speed-up} and finds better approximations than the current state-of-the-art. First, we map the optimization problem into a boolean circuit made of specially designed, {\it self-organizing} logic gates, which can be built with (non-quantum) electronic components~\cite{DMM2}; the equilibrium points of the circuit represent the approximation to the problem at hand. Then, we solve its associated {\it non-linear} ordinary differential equations numerically, towards the equilibrium points. We demonstrate this exponential gain by comparing a sequential MatLab implementation of our solver with the winners of the 2016 Max-SAT competition on a variety of hard optimization instances.  We show empirical evidence that our solver scales {\it linearly} with the size of the problem, both in time and memory, and argue that this property derives from the {\it collective} behavior of the simulated physical circuit. Our approach can be applied to other types of optimization problems and the results presented here have far-reaching consequences in many fields.}

In real-life applications it is common to encounter problems where one needs to find the best solution within a vast set of possible solutions. These {\it optimization problems} are routinely faced in many commercial segments, including transportation, goods delivery, software packages or hardware upgrades, network traffic and congestion management, and circuit design, to name just a few ~\cite{Optimization_book,Optimization_book_intro}.  
Many of these problems can be easily mapped into {\it combinatorial optimization problems}, namely they can be written as boolean formulas with many constraints (clauses) among different variables, (either negated or not i.e., literals), with the constraints themselves related by some logical proposition~\cite{Optimization_book}.  

It is typical to write the boolean formulas as {\it conjunctions} (the logical ANDs, also represented by the symbol $\wedge$) of {\it disjunctions} (the logical ORs, represented by the symbol $\vee$), in the so called {\it conjunctive normal form} (CNF). The CNF representation is universal in that any boolean formula can be written in this form~\cite{complexity_bible}.

A simple example of a CNF formula $\varphi(x)$ is
\begin{eqnarray}
\varphi(x)&=&(\lnot x_{1}\mathbf{\vee}x_{2})\mathbf{\wedge}(\lnot
x_{2}\mathbf{\vee}\lnot x_{3}\mathbf{\vee}x_{4})\mathbf{\wedge}\nonumber \\
&&(x_{1}%
\mathbf{\vee}\lnot x_{2}\mathbf{\vee}x_{3}\mathbf{\vee}\lnot x_{4}%
)\mathbf{\wedge}\nonumber \\
&&(\lnot x_{1}\mathbf{\vee}x_{4})\mathbf{\wedge}(x_{1}%
\mathbf{\vee}x_{2}\mathbf{\vee}\lnot x_{4})\label{exampleCNF}\nonumber
\end{eqnarray}
in which we have four variables, $x_j$, with $j=1,2,3,4$, five clauses, and fourteen literals (the symbol $\lnot$ indicates negation). 
The problem is then to find an assignment satisfying the maximum number of clauses, i.e., in which as many clauses as possible have at least one literal that is true. Such a clause is then said to be satisfied, otherwise it is unsatisfied~\cite{complexity_bible}, and the problem itself is known as Max-SAT (maximum satisfiability). 

A Max-SAT problem whose CNF representation has exactly $k$ literals ($k\geq 2$) per clause, is called Max-E$k$SAT.  
Max-E$k$SAT is a ubiquitous optimization problem with widespread industrial applications. We will focus on its solution as a test bed in the main text, and refer the reader to the Supplemental Information for other optimization problems, including weighted SAT.

Due to its NP-hard nature, complete algorithms that attempt to solve Max-E$k$SAT quickly become  unfeasible for large problems. Much research has instead focused on incomplete solvers that perform a stochastic local search, by generating an initial assignment, and iteratively improving upon it.  This approach has proven effective at approximating and sometimes solving large instances of
SAT and other problems.  For instance, in recent Max-SAT competitions~\cite{MAXSAT_competition}, incomplete solvers outpace complete solvers by two orders of magnitude on random and crafted benchmarks, and perform similarly on industrial problems. However, they too suffer from the same exponential time dependence as complete solvers for sufficiently large or hard instances \cite{gomes2008satisfiability,kautz2009incomplete,heuristics_book}.

In the worst cases, it has been shown using probabilistically checkable proofs~\cite{Feige1998} that many classes of combinatorial optimization problems (including the Max-E$k$SAT) have an {\it inapproximability gap}. This means that no algorithm can overcome, in polynomial time, a fraction of the optimal solution, unless NP=P~\cite{Feige1998,Hastad2001}. In other words, for heuristics to improve on their approximation would require exponentially increasing time. For example, for the Max-E3SAT it has been proved that, if NP$\neq$P, then there is no algorithm that can give an approximation better than 7/8 of the optimal number of satisfied
clauses~\cite{Hastad2001}.

%More formally, the Max-E$k$SAT belongs to the APX-complete class, i.e., the class of optimization problems that, assuming NP$\neq$P, admits a polynomial-time approximation algorithm with an approximation ratio bounded by a constant, and it admits a polynomial mapping to any other NP optimization problem. Therefore, as anticipated, an APX problem has a ``gap of approximability'', meaning that one can approximate the problem in polynomial time with respect to $|x|$ (the size of the input) only up to a limit. In order to improve beyond this approximation a time exponentially growing with $|x|$ is required. 

In this work, we consider instead a radically different {\it non-combinatorial} approach to hard optimization
problems. Our approach is based on the {\it simulation} of {\it digital memcomputing machines} (DMMs)~\cite{DMM2,UMM} discussed in Methods. Their practical realization can be accomplished using standard circuit elements and those with memory (time non-locality, hence the name ``memcomputing''~\cite{diventra13a}). 

Time non-locality allows us to build logic gates that {\it self-organize} into their logical proposition, {\it irrespective} of whether the signal comes from the traditional input or output~\cite{DMM2}. We call them {\it self-organizing logic gates} (SOLGs), and circuits built out of them, {\it self-organizing 
	logic circuits} (SOLCs). Our approach then follows these steps. 
\begin{figure}[t]
	%\begin{center}
	\includegraphics[width=1\columnwidth]{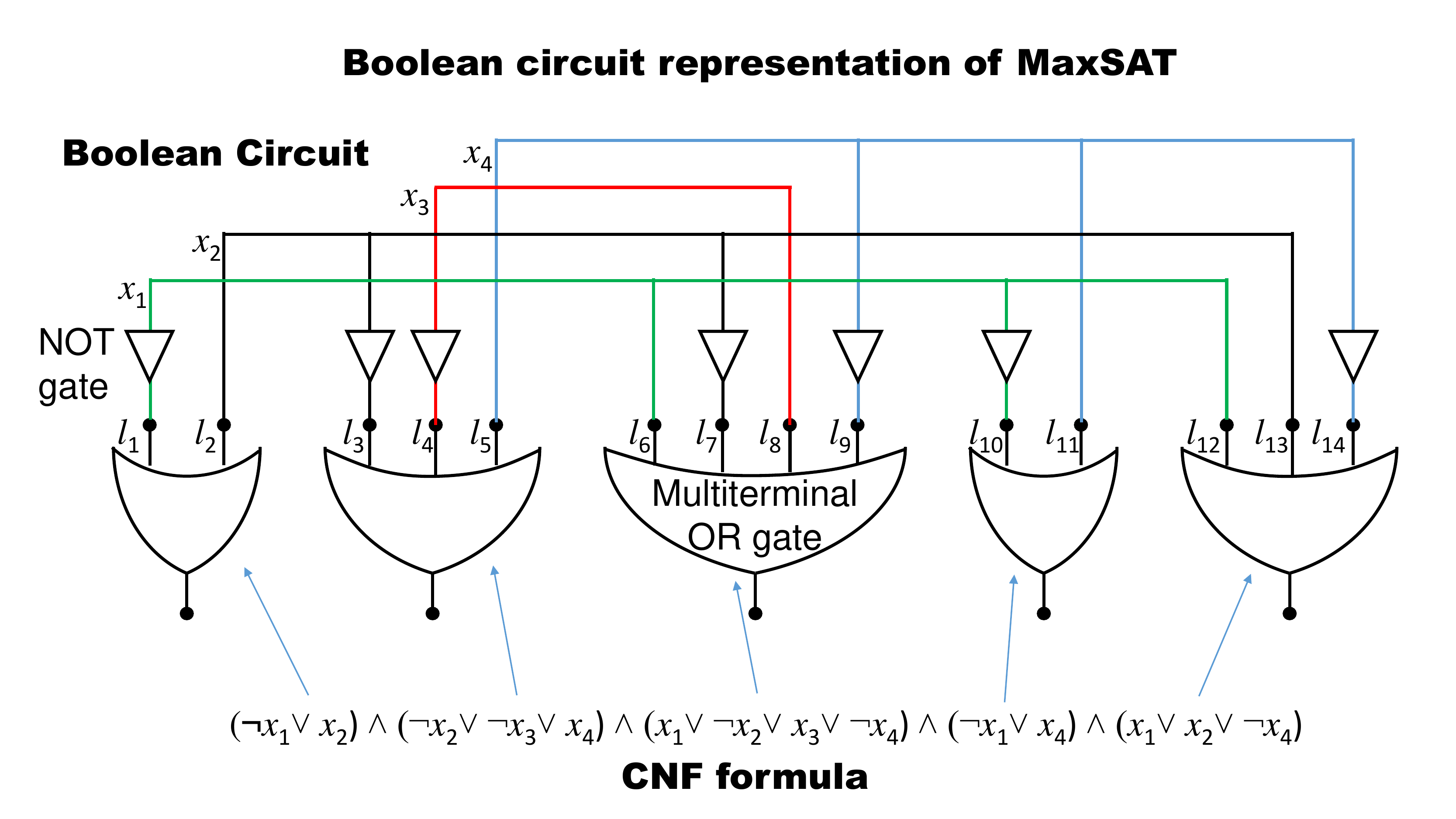}
	%\end{center}
	\caption{Example of the mapping between a boolean satisfiability formula in conjunctive
		normal form and a boolean circuit made of multi-terminal OR and NOT gates. Each
		clause of the SAT formula is mapped into an OR with as many terminals as the
		literals in the clause (the satisfiability of this multi-terminal OR requires that at least one terminal has a truth value of 1). The global optimum of the SAT formula, i.e., the maximum 
		number of satisfied clauses, corresponds to the maximum number of OR gates with 
		output one. This boolean circuit is then transformed into a self-organizing logic circuit by
		substituting each standard boolean gate with a self-organizing logic gate~\cite{DMM2}, and each OR output is fed with a DC voltage 
		generator representing the logic value of 1.}%
	\label{Fig_1}%
\end{figure}

1) We first construct the boolean circuit that represents the problem at hand (e.g., the Max-E$k$SAT of Fig. \ref{Fig_1}). 

2) We replace the traditional (uni-directional) boolean gates of this boolean circuit with SOLGs. 

3) We feed the appropriate terminals with the required output of the problem (e.g., the logical 1 if we are interested in checking its satisfiability). 

4) Finally, the electronic circuit built out of these SOLGs can be described by {\it non-linear} ordinary differential equations, which can be solved to find the equilibrium (steady-state) points. These equilibria represent the approximation to the optimization problem~\cite{DMM2}. 

The procedure of how we transform a combinatorial optimization problem into an electronic circuit, as well as a sketch of its numerical solution is discussed further in the Methods section. The important point to note is that SOLGs and SOLCs manifest {\it long-range order} due to the presence of instantons~\cite{topo}. Instantons connect topologically inequivalent critical points in the phase space, hence generating {\it non-locality} in the system. This translates into a 
{\it collective} dynamical behavior that allows gates at an {\it arbitrary} distance to correlate very efficiently so that, when a terminal of one gate needs to change its truth value to satisfy that gate's logical proposition, a terminal at any other gate may provide the correct truth assignment while satisfying its own logical proposition. As we will explain later, this is the key feature that allows these memcomputing machines to solve complex problems efficiently, without the need to explore a vast space of possibilities, as standard combinatorial approaches would do.

This radical change of perspective manifests its power already in comparing simulations of DMMs with those performed by the winners of the 2016 Max-SAT competition~\cite{MAXSAT_competition} on the competition benchmarks. When run on similar hardware, our solver, which we named Falcon \cite{DMM2,13_amoeba,DCRAM}, performs orders of magnitude faster than the winners in the incomplete track of the competition, and in some cases it finds the solution when the best solvers did not.  

Since a direct comparison is difficult across hardware and implementations (our solver is written in MatLab which is notoriously inefficient compared with the compiled languages of the competition solvers), we have presented these results in Figs.~\ref{Fig_SI_1}, \ref{Fig_SI_2}, \ref{Fig_SI_3}, and \ref{Fig_SI_4} of the Supplemental Information. Nevertheless, these tests already provide strong indication of the advantages of our approach using digital memcomputing machines over traditional combinatorial optimization.

However, in order to form a direct comparison and more clearly show the {\it exponential speed-up} of our approach, we have crafted three Max-SAT problems with increasing levels of difficulty. We then compared our memcomputing solver against two of the best solvers of the 2016 Max-SAT competition, (CCLS~\cite{CCLS} and DeciLS~\cite{decils} --a new version of CnC-LS--, kindly provided by their developers) which are specifically designed to solve these types of problems, but employing very different solution strategies. (Note that we expect other types of algorithms, e.g., those based on message passing, to show the same behavior as these two local solvers for the balanced and constrained Max-XORSAT instances we consider in this work~\cite{Mezard}.)

Random 3-SAT instances may be generated by selecting 3 variables out of $n$, joining them in a 3-SAT clause where each is randomly negated, and then repeating this for the desired number of clauses $M$.  These instances are known to undergo a SAT/UNSAT transition when the ratio of clauses to variables, $M/n = \rho$ (hereafter the ``density''), crosses the critical value $\rho_c \approx 4.3$~\cite{mitchell1992hard,kirkpatrick1994critical}.  Exponential time is required to demonstrate that an instance is UNSAT~\cite{cocco2006} and thus must also be required to solve the corresponding Max-SAT, offering a simple way to generate benchmarks.

However, the difficulty of computing approximations for these instances varies widely.  This can be partially attributed to the fluctuations in variable occurrences and their negations~\cite{barthel2002} leading to `fields' which point towards the optima.  More balanced instances may be produced by starting with a Random-XORSAT instance (also called hyperSAT~\cite{ricci2001simplest} ) i.e., a set of boolean formulas defined by the XOR of boolean variables (the XOR symbol is $\oplus$) and converting it to a Max-SAT instance.

Each XORSAT clause may be converted to a block of four SAT clauses, e.g., $x\oplus y\oplus z=1\rightarrow(x\mathbf{\vee}y\mathbf{\vee}z)\mathbf{\wedge}(x\mathbf{\vee
}\lnot y\mathbf{\vee}\lnot z)\mathbf{\wedge}(\lnot x\mathbf{\vee}\lnot y\mathbf{\vee}z)\mathbf{\wedge}(\lnot
x\mathbf{\vee}y\mathbf{\vee}\lnot z)$, in which a variable and its negation appear symmetrically.  The special structure of XORSAT gives rise to a global algorithm when the instance is satisfiable, allowing for a solution in polynomial time using Gaussian elimination~\cite{ricci2001simplest}.  However, when unsatisfiable, occurring for $\rho > 4\cdot 0.918 \approx 3.7$, this same structure makes these problems very difficult for local search solvers~\cite{cocco2006, jia2004}.

A basic understanding of this difficulty can be obtained by considering that changing a variable assignment  affects positively (namely contributes a true literal to) the same number of clauses as those affected negatively (where the literal is false), because of the balanced occurences of the variables. Therefore, for any combinatorial approach, when a certain amount of satisfied clauses is reached, any further improvement requires many {\it simultaneous} variable flips, which is a {\it non-local} type of assignment. 
In other words, the distance between two assignments at successive approximations becomes of the same order of the
input length $|x|$. This means that going from an assignment $x$ to a better one $y$, if they have a distance $d(x,y)=\sum_{j}(x_{j}-y_{j})^{2}=O(|x|)$, would require checking $O(2^{d(x,y)})$ variable flips, which is a number of configurations that is exponential
with respect to the distance $d(x,y)$. (The actual calculation requires the enumeration of all possible flips of $1, 2, ..., d(x,y)$ literals because the distance $d(x,y)$ is not known {\it a priori}. Hence, the actual number of flips is 
${\sum_{k=0}^{d(x,y)}} \binom{|x|}{k}\geq2^{d(x,y)}$.)
	
While more difficult, these instances also display wide variation in resolution time. In order to obtain instances of more predictable difficulty, we impose a further constraint requiring all variables to appear the same number of times (or as near as possible while remaining consistent with the number of clauses $M = \rho N$), i.e., the variable occurrences are distributed as a $\delta$-function. This variant is expected to be harder than the previous one because of the additional balance induced by the variable distribution, and our results indicate that they display much lower variability in their difficulty. 

In the following, we will call ``random-Max-E3SAT'' a Max-E3SAT completely generated at random. This will be used as an ``easy'' problem to test the performance of all solvers. We refer to ``hyper-Max-E3SAT'' as the Max-E3SAT generated from a random Max-E3XOR, and finally to ``delta-Max-E3SAT'' as a problem generated by the Max-E3XOR with $\delta$-function distribution of variables. 

While the balanced structure of Max-XORSAT poses a challenge to local search algorithms (or message-passing-based ones), our memcomputing solver easily overcomes these limits because, due to the collective (instantonic) behavior of the circuit, the dynamics evolve towards deep minima very close to the global optimum (see also Methods). The reason is that, as already anticipated, the collective state of the machine allows simultaneous, {\it non-local} change of literals belonging to gates arbitrarily far from each other~\cite{topo}. This change is consistent with the physics and the topology of the memcomputing circuit that naturally drive the system towards the maximum number of satisfied SOLGs, without recourse to any combinatorial selection scheme. 

The optimum for all problems can be estimated using an ensemble of small instances for which it is easier to find a fairly good approximation. For example, instances of about 300 variables and density (clauses/variables) of $\rho=5$ provide a good indication of the global optimum in terms of percentage of unsatisfied clauses. We found that for the random-Max-E3SAT the optimum is expected at about $0.4\%$ of unsatisfied clauses, while for both the hyper- and delta-Max-E3SAT this value is about $1.3\%$. The difference between these values is not surprising. As mentioned previously, it is well known that for the latter two problems the transition from satisfiable to unsatisfiable is around a density of $\rho\approx3.7$, while for random-Max-E3SAT it is around $\rho\approx 4.3$. We have then chosen the same density of $\rho=5$ for the random-, hyper- and delta-Max-E3SAT. 
\begin{figure*}[t]
	%	\begin{center}
			\includegraphics[width=1.8\columnwidth]{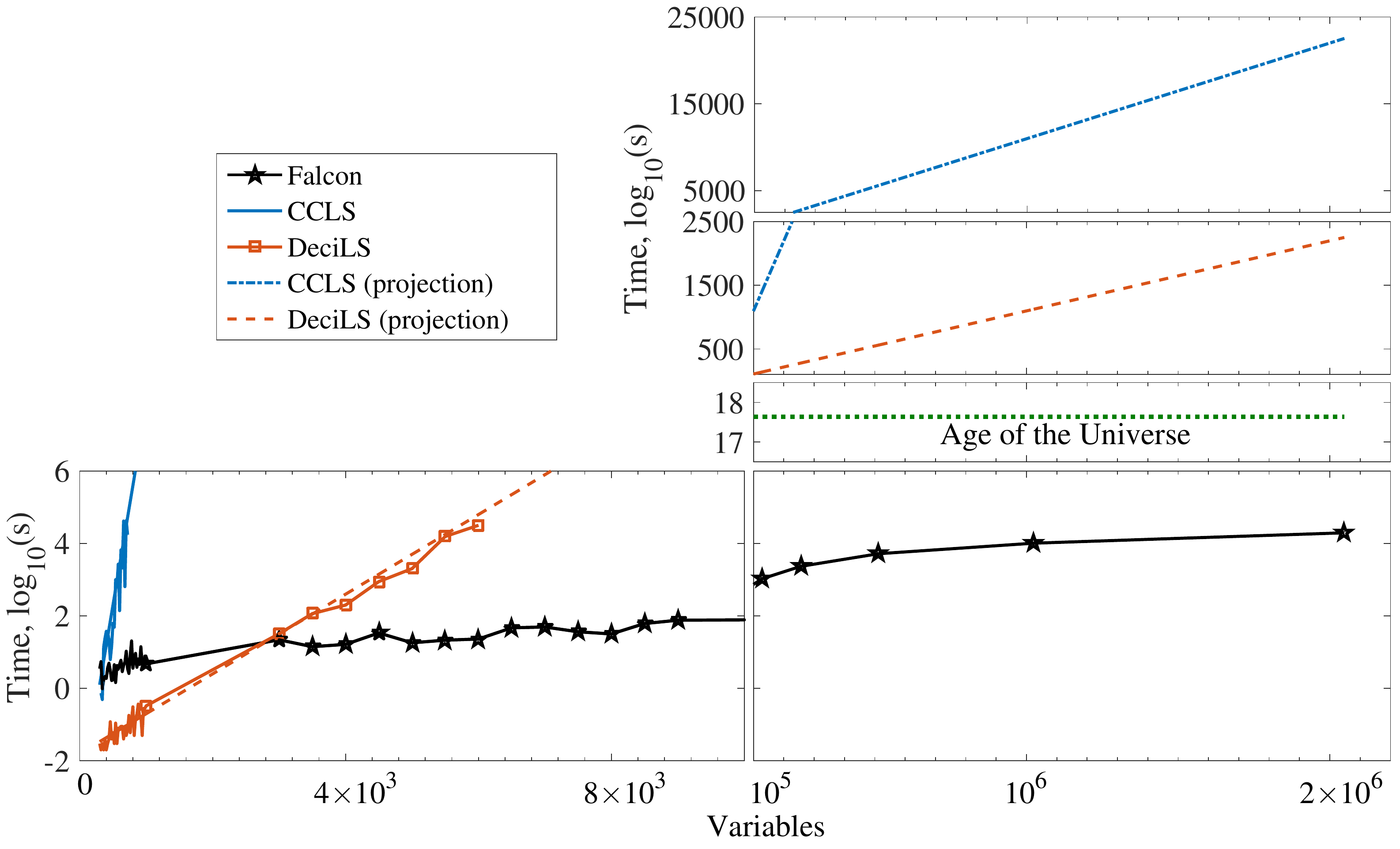}	
	%	\end{center}
	\caption{Simulation time comparison between incomplete solvers CCLS and DeciLS against our solver, Falcon, for the balanced and constrained delta-Max-E3SAT. A threshold of $1.5\%$ of unsatisfiable clauses has been set. We have then tested how long CCLS, DeciLS and our solver Falcon take to overcome this limit with increasing number of variables. All calculations have been performed on a single thread of an Intel Xeon E5-2680 v3 with 128 Gb DRAM shared on 24 threads. The local solvers require an exponentially increasing time to reach that limit already visible at a few hundred variables for the CCLS and a few thousands for the DeciLS. Our solver has been tested up to $2\times 10^6$ variables, and required order of $10^4$ seconds for that maximum number of variables. We show also the estimate of time that would have been required these local solvers to run up to $2\times 10^6$ variables. The estimated time (dashed and dashed-dotted lines) has been calculated using a linear regression of the $\log_{10}($time$)$ versus the number of variables.}%
	\label{Fig_2}%
\end{figure*}

In order to prove the superior efficiency of our non-combinatorial approach for this class of hard problems, we have evaluated their scaling properties up to $2\times10^{6}$ variables (while keeping the density constant). We recall that the simulations of DMMs have been done using a MatLab code, while CCLS and DeciLS are compiled codes. Therefore, the level of optimization is expected to be higher in the compiled codes, making a direct performance comparison harder, although for large problem sizes, our solver has much better performance compared to CCLS and DeciLS. Nevertheless, we are more interested in the scaling of the approximation time. Specifically, for hard cases where incomplete solvers diverge exponentially in time, our solver diverges {\it linearly}. This is the most important test and the central result of our paper. It is shown in Fig.~\ref{Fig_2}.

The hard inapproximability limit and its exponential nature for both the combinatorial heuristics CCLS and DeciLS is clearly visible in Fig. \ref{Fig_2}, where we have set a threshold of $1.5\%$ of unsatisfiable clauses for the delta-Max-E3SAT. We have then tested how long CCLS, DeciLS and our solver Falcon take to overcome this limit with increasing number of clauses. All calculations have been done on a single core of an Intel Xeon E5-2680 v3. 

The exponential blowup of CCLS and DeciLS is already evident for small instances of the problem, while our non-combinatorial approach performs {\it linearly}, in both time and memory, for {\it any} number of variables we have tested so far. In fact, we have tested 
our solver up to $2\times 10^6$ variables, requiring $\sim 10^4$ seconds to reach the target $1.5\%$ threshold. The heuristic solvers, if they could run up to the same number of variables would require, in the best case, about $\sim 10^{2500}$ seconds, which is $\sim 10^{2480}$ times the estimated age of the Universe.  
\begin{figure}[t]
	\centering
	\begin{overpic}[width=\columnwidth]{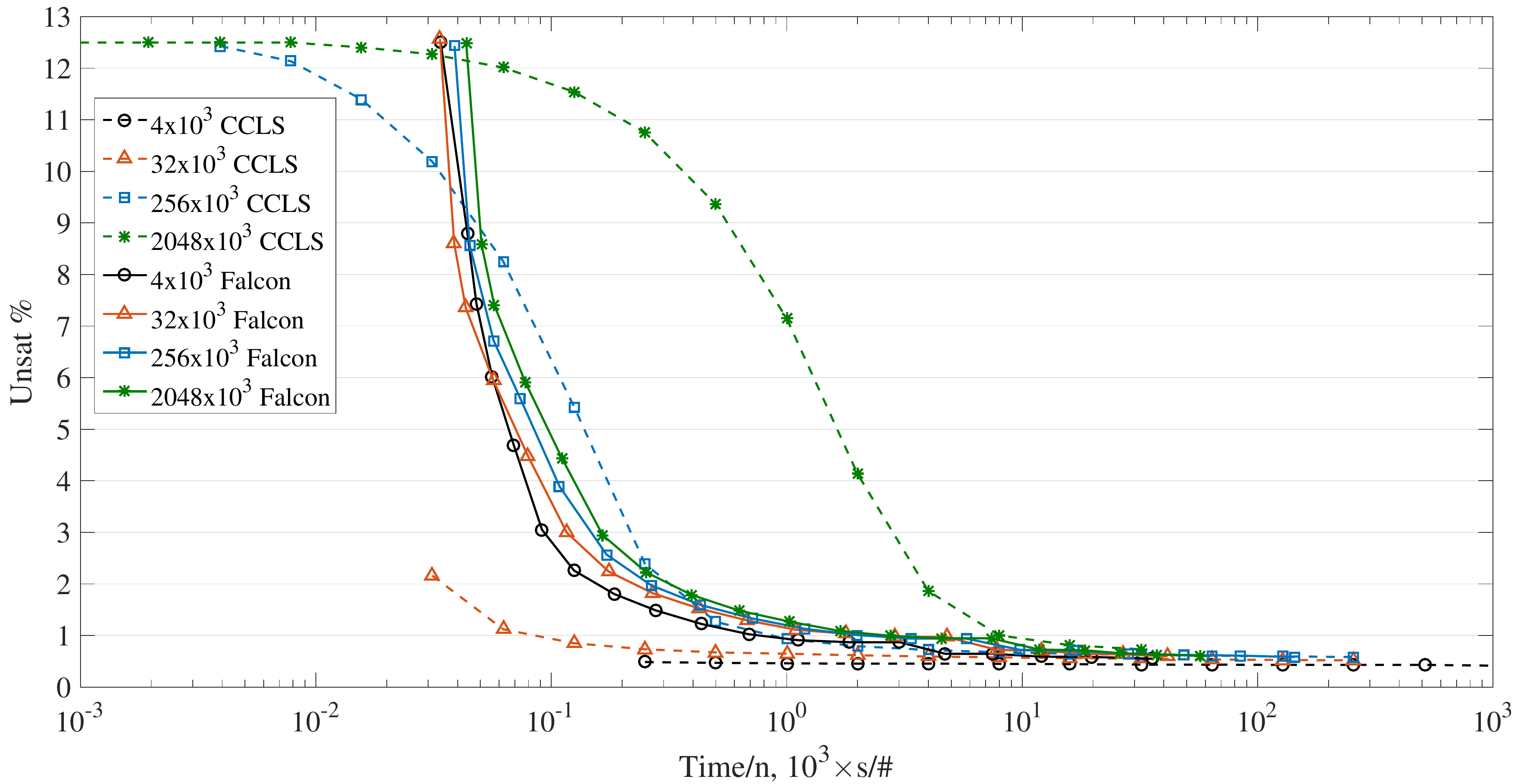}
		\put (5,52) {(a)}
	\end{overpic}
	\par\vspace{.2cm}
	\begin{overpic}[width=\columnwidth]{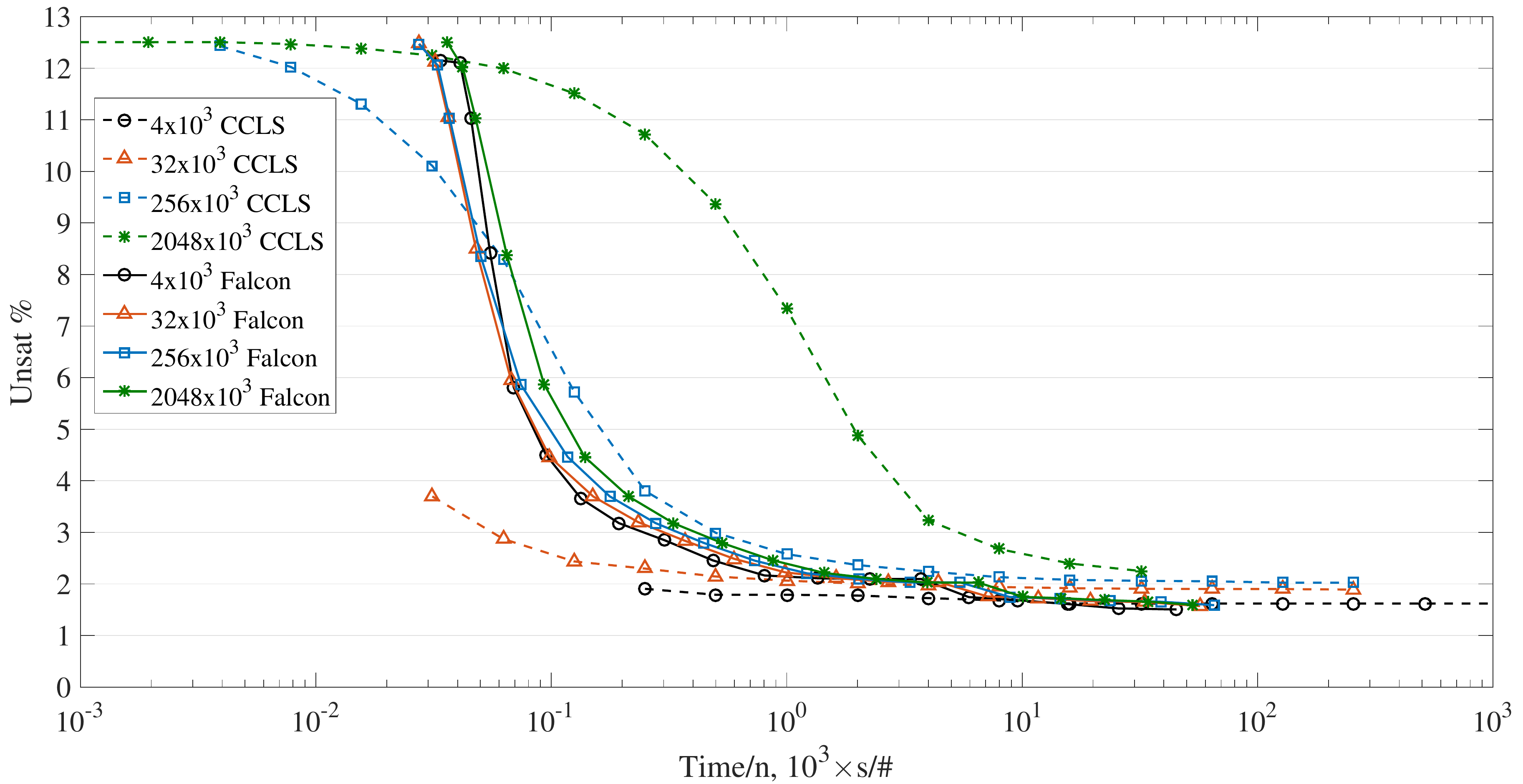}
		\put (5,52) {(b)}
	\end{overpic}
	\par\vspace{.2cm}
	\begin{overpic}[width=\columnwidth]{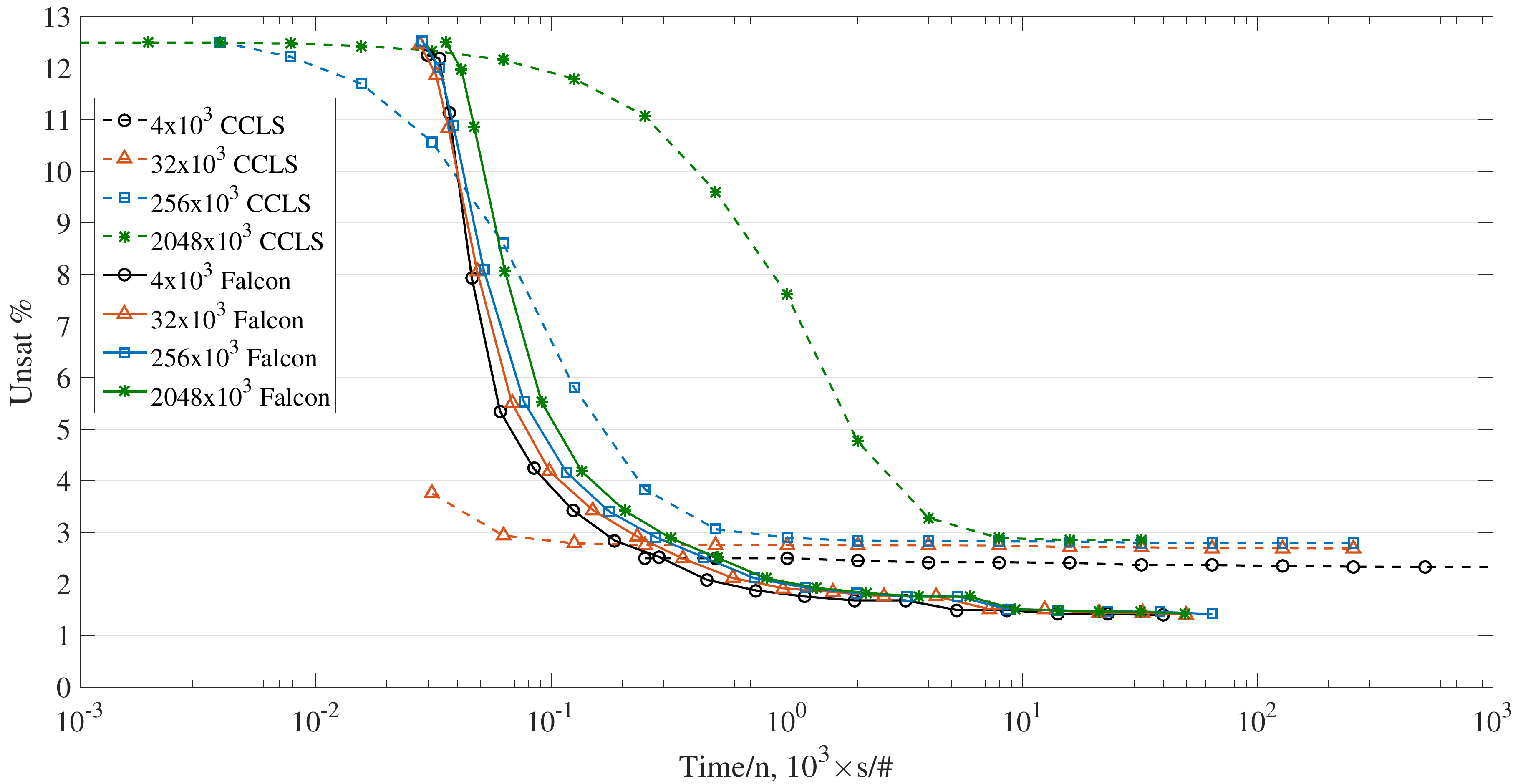}
		\put (5,52) {(c)}
	\end{overpic}
	\caption{Comparison between the incomplete solver CCLS versus our non-combinatorial solver, Falcon, for (a) Random-Max-E3SAT, (b) hyper-Max-E3SAT, (c) delta-Max-E3SAT. In these plots the percentage of unsatisfied clauses versus the time normalized with respect to the number of 
		variables is shown to highlight the linear scaling of our solver.  All 
		calculations have been performed on a single thread of an Intel Xeon E5-2680 v3 with 128 Gb DRAM shared on 24 treads.}%
	\label{Fig_3}%
\end{figure}

To better highlight the {\it linear} scaling of our solver, we compare it in Fig.~\ref{Fig_3} with CCLS (qualitatively, all other incomplete solvers should perform 
similarly).  
Each plot of Fig.~\ref{Fig_3} displays the percentage of unsatisfied clauses versus time, normalized with respect to the number of variables $n$. Clearly, linear scaling for these hard problems is a very desirable feature, and very difficult to achieve with combinatorial approaches. 
However, the reason for such linear scaling is subtle. 

\begin{figure}[ptb]
		\centering
	\begin{overpic}[width=\columnwidth]{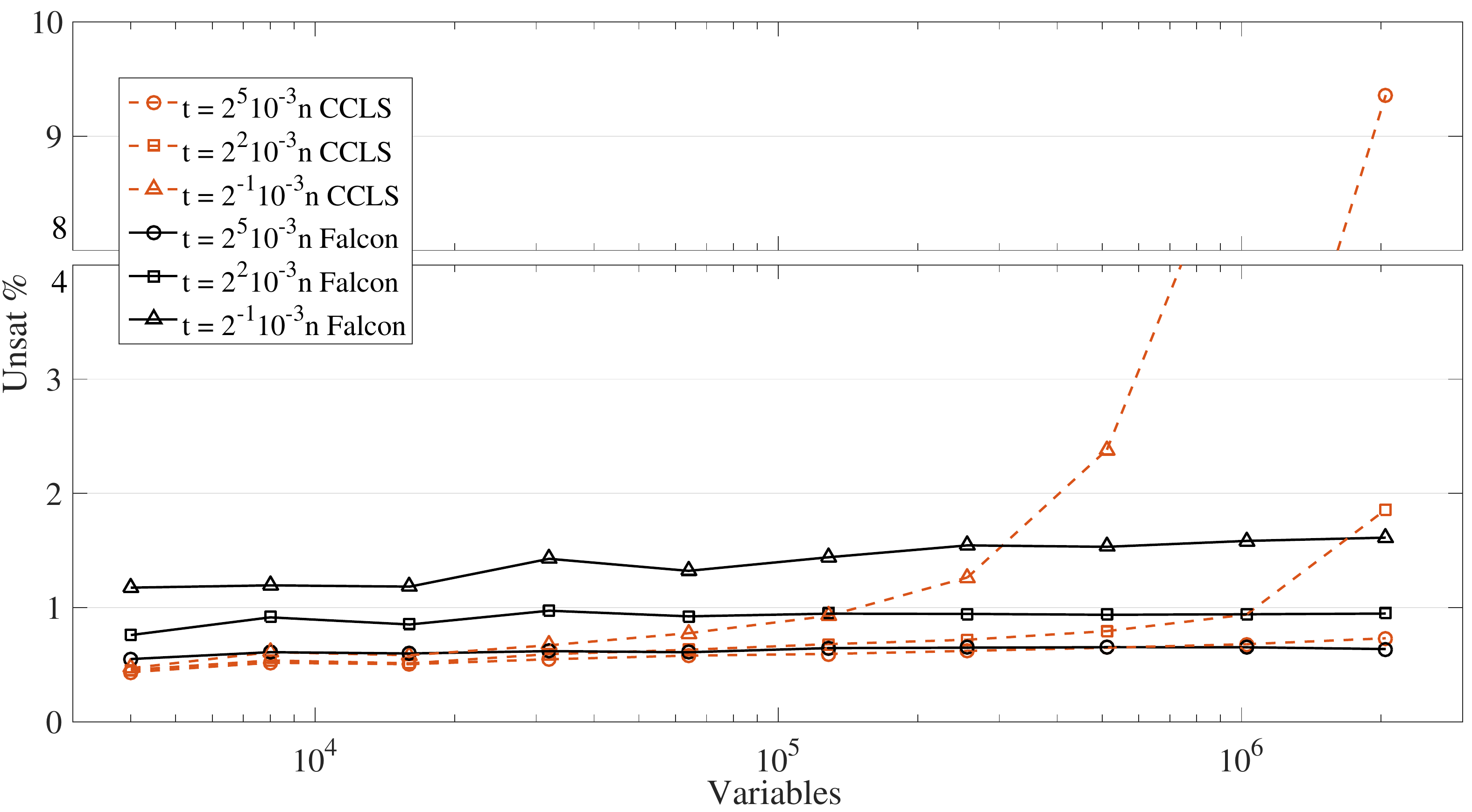}
		\put (5,52) {(a)}
	\end{overpic}
	\par\vspace{.2cm}
	\begin{overpic}[width=\columnwidth]{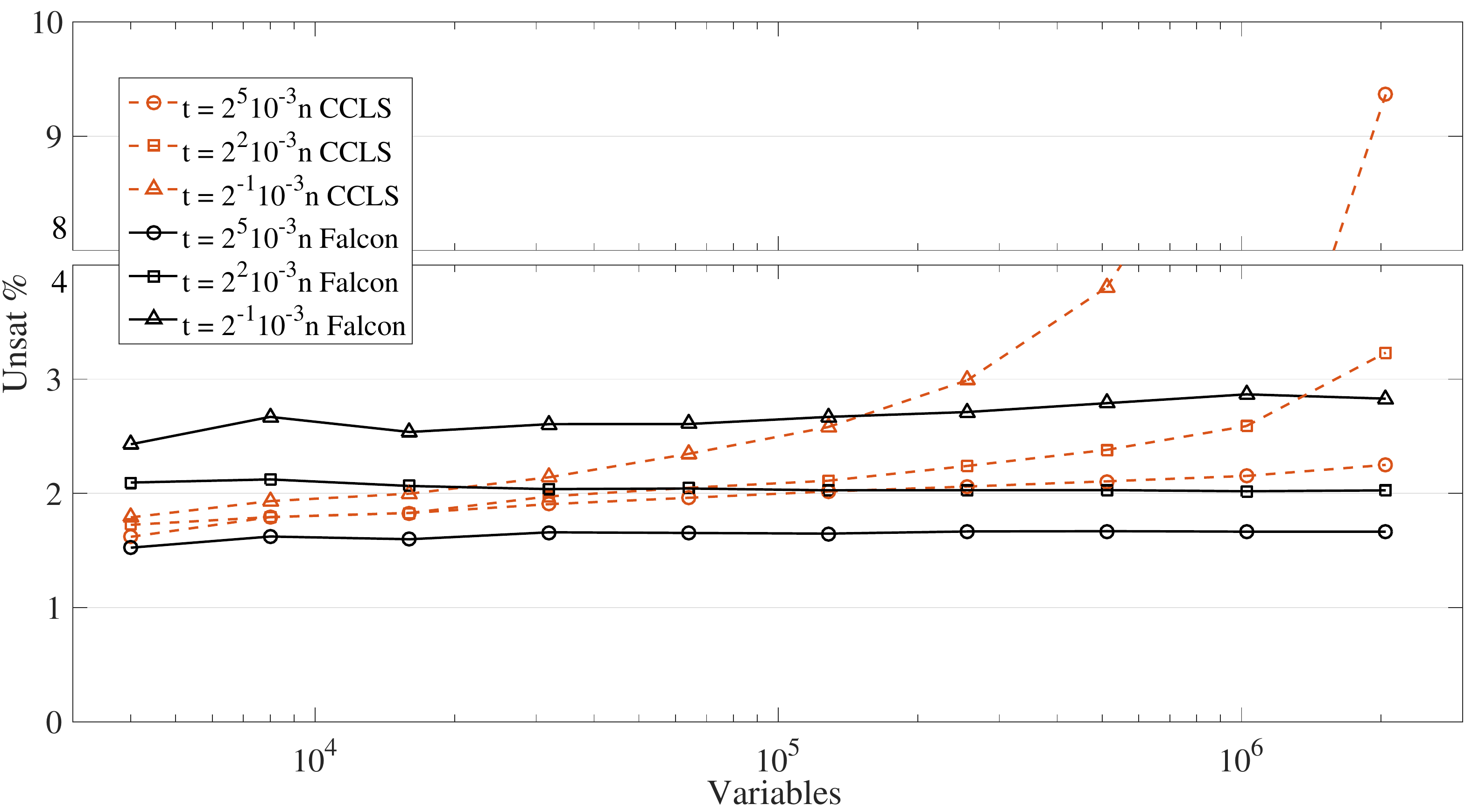}
		\put (5,52) {(b)}
	\end{overpic}
	\par\vspace{.2cm}
	\begin{overpic}[width=\columnwidth]{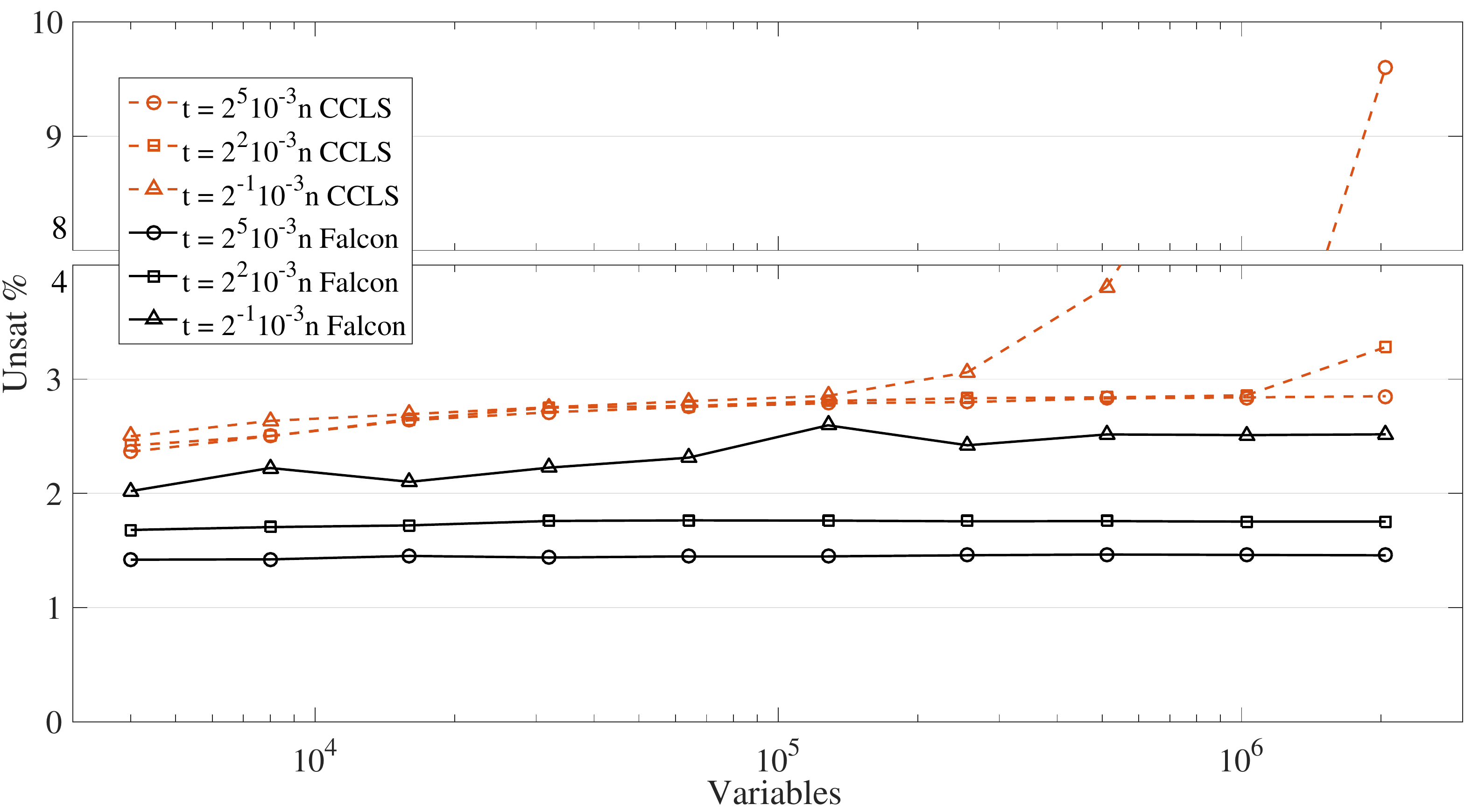}
		\put (5,52) {(c)}
	\end{overpic}
	\caption{Comparison between the CCLS solver versus our solver, Falcon,  for (a) random-Max-E3SAT, (b) hyper-Max-E3SAT, (c) delta-Max-E3SAT. 
		In these plots the percentage of unsatisfied clauses versus the number of variables is shown. Different curves are for different simulation time outs (in seconds) following the relation $t_{out} = kn$ with $n=|x|$, and $k$ an integer given in the legend. All 
		calculations have been performed on a single thread of an Intel Xeon E5-2680 v3 with 128 Gb DRAM shared on 24 treads.}%
	\label{Fig_M_1}%
\end{figure}

Regarding {\it memory}, since we simulate (integrate) differential equations in time, and the circuit scales linearly with the number of literals, the linear scaling in memory requirements of our simulations is easy to understand (see also Methods). On the other hand, linear scaling in simulation {\it time} implies {\it constant} scaling, namely {\it independent} of the problem size, when we look at the ``machine time'', which is the number of (differential equation discretized time) steps for the simulation to reach equilibrium. The reason for this unexpected machine-time constant scaling can be found again in the long-range order of the dynamics of the system~\cite{topo} (see also the Methods section). As we have shown analytically in Ref.~\cite{topo} using topological field theory, this long-range order leads to non-decreasing spatial (and temporal) correlations in memcomputing machines (see Methods for further discussion). 
In fact, Fig.~\ref{Fig_3} clearly shows that self-organizing logic circuits relax close to the predicted global minimum,
while the CCLS does so only for the (``easy'') random-Max-E3SAT. This is further illustrated in Fig.~\ref{Fig_M_1} of the Methods section 
for random-, hyper-,  and delta-Max-E3SAT. 

In conclusion, we have shown empirical evidence that a {\it non-combinatorial} approach --based on the simulation of digital memcomputing machines-- to the solution of hard combinatorial optimization problems outperforms {\it exponentially} heuristics specifically designed to solve such problems. In 
particular, with our approach we were able to find far better approximations to hard instances with millions of variables in a few hours on a single core, with {\it linear scaling} both in time and memory of the processor. For the same sizes, winners of the 2016 Max-SAT competition would require several orders of magnitude more 
than the age of the Universe to find the same approximations.
Of course, these numerical results are not intended to prove that there are polynomial solutions to NP-hard problems. Rather, they show that physics-inspired 
approaches can help tremendously in solving some of the most complex problems faced in academia and industry. We thus hope that this work will motivate further research along these lines. \\

{\it Acknowledgments --} We sincerely thank Dr. Shaowei Cai for providing us with the binary compiled codes CCLS and DeciLS. We also thank Haik Manukian and Robert Sinkovits for helpful discussions. M.D. and F.L.T. acknowledge partial support from the Center for Memory Recording Research at UCSD. M.D. and F.S. acknowledge partial support from MemComputing, Inc. All calculations reported here have been performed by one of us (P.C.) on a single processor of the Comet cluster of the San Diego Supercomputer Center, which is an NSF resource funded under award \#1341698.

\begin{figure}[ptb]
	\centering
	\begin{overpic}[width=\columnwidth]{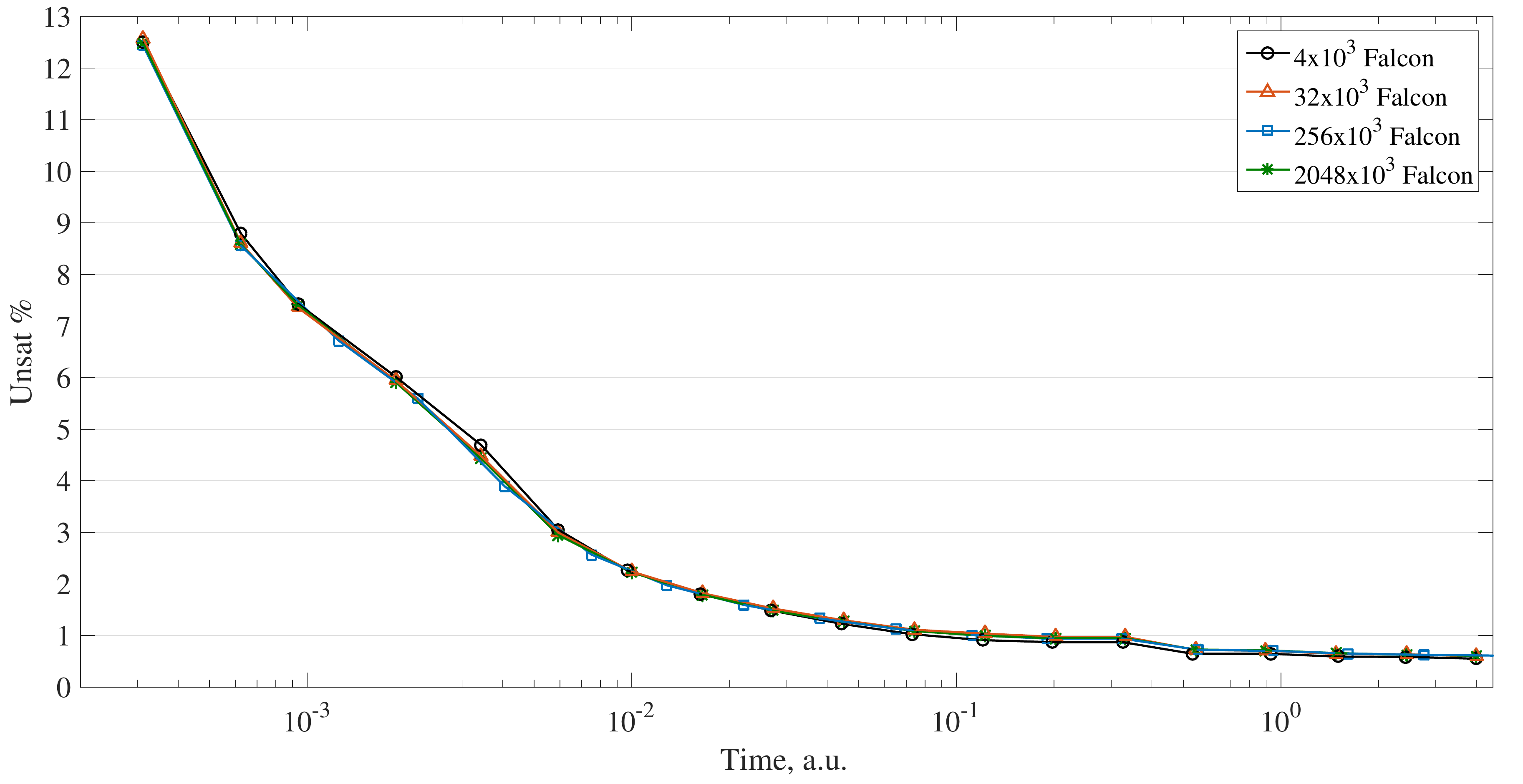}
		\put (5,52) {(a)}
	\end{overpic}
	\par\vspace{.2cm}
	\begin{overpic}[width=\columnwidth]{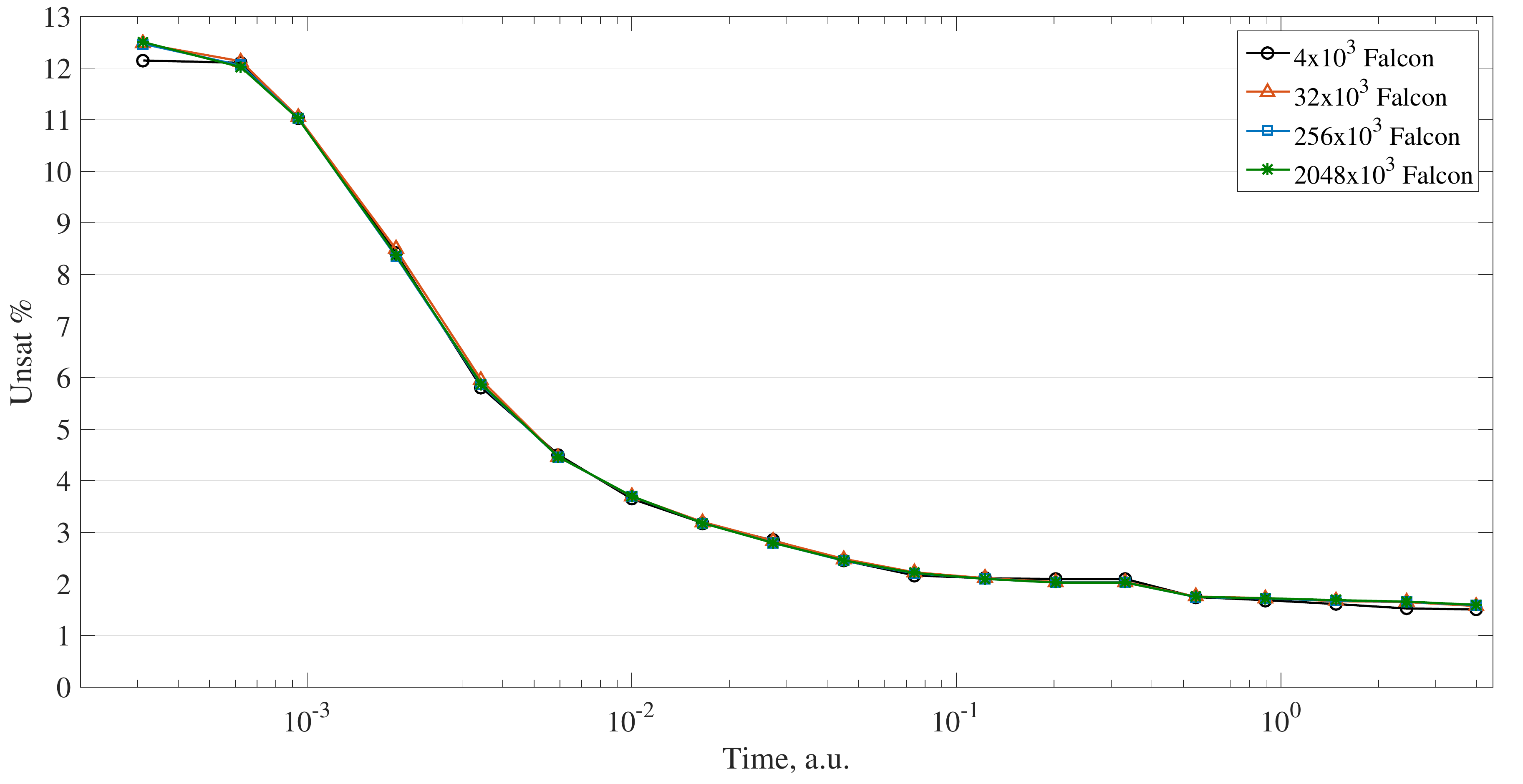}
		\put (5,52) {(b)}
	\end{overpic}
	\par\vspace{.2cm}
	\begin{overpic}[width=\columnwidth]{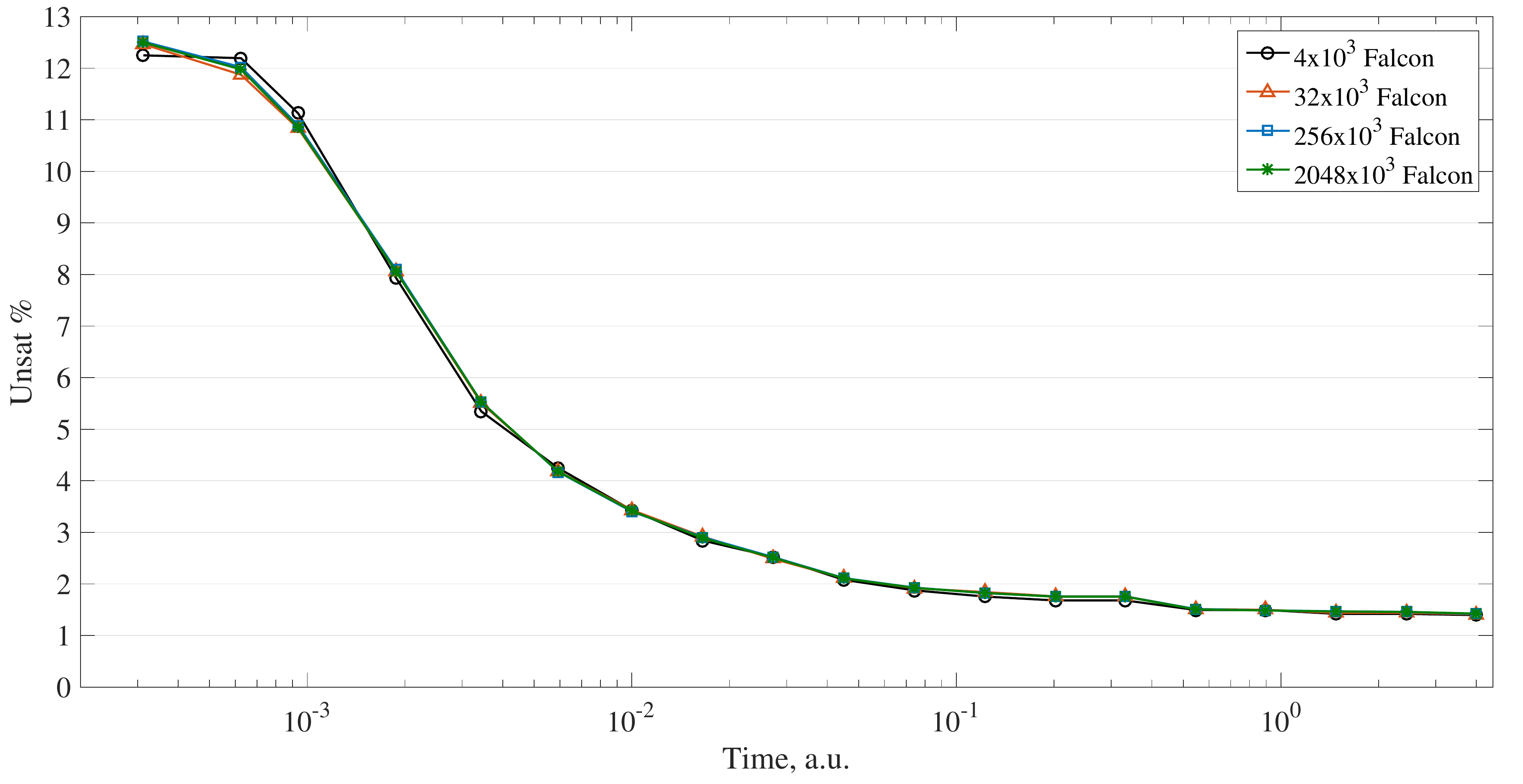}
		\put (5,52) {(c)}
	\end{overpic}
	\caption{Percentage of unsatisfied clauses versus the machine time (i.e., simulated time steps) is shown to highlight the linear scaling of our solver, Falcon,  for (a) random-Max-E3SAT, (b) hyper-Max-E3SAT, (c) delta-Max-E3SAT. All 
		calculations have been performed on a single thread of an Intel Xeon E5-2680 v3 with 128 Gb DRAM shared on 24 treads.}%
	\label{Fig_M_2}%
\end{figure}

Apart from the instances freely available from the 2016 Max-SAT competition~\cite{MAXSAT_competition}, the authors would be delighted to provide, upon request, all instances of the constrained delta-Max-E3SAT used to generate Fig.~\ref{Fig_2}, and those related to Figs.~\ref{Fig_3},~\ref{Fig_M_1}, and~\ref{Fig_M_2}.\\

{\bf \large Methods}

The non-combinatorial approach we discuss here is based on the concept of Universal Memcomputing Machines (UMMs)~\cite{UMM} introduced by two of us (F.T. and M.D.). 
UMMs are a class of computing machines composed of interconnected memory units. The {\it topology} of such network is chosen to solve the specific problem at hand. UMMs use the {\it collective} state of the interconnected memory units to perform computation~\cite{DMM2,traversaNP}, so they can take advantage of long-range correlations that can significantly boost the efficiency of the computation~\cite{DMM2,topo}. If the input and output of UMMs can be mapped into strings of integers, belonging to a limited subset of $\mathbb{N}$, we obtain the digital (hence scalable) version of UMMs (DMMs)~\cite{DMM2}. In particular, we consider DMMs whose input and output can be mapped into $\mathbb{Z}_{2}$.

A possible, practical realization of DMMs are self-organizing logic circuits (SOLCs) composed of SOLGs~\cite{DMM2}. SOLGs are
logic gates that can accept inputs from {\it any} terminals, and self-organize their internal state to satisfy their logic relations. For example, a self-organizing OR (SO-OR) is a 3-terminal gate whose internal machinery drives the terminal states to satisfy the relation $x_{o}=x_{1}\vee x_{2}$, where $x_{o}$ is the state of the conventional output terminal, and $x_{1}$, $x_{2}$ are the states of the conventional input terminals. Therefore, unlike conventional logic gates, the SO-OR can be fed also at the output terminal. If we set $x_{o}$ to some state, the SO-OR then will self-organize to give consistent states $x_{1}$ and $x_{2}$.

We can use SOLCs to solve combinatorial problems by expressing the problem in boolean format and then mapping the latter onto logic circuits. As a relevant example for this work, we can take the Max-SAT problem written in
CNF. When we transform the SAT into a boolean circuit we have multi-terminal OR gates connected together in order to represent a logic formula (see Fig. \ref{Fig_1}). Hence, we can substitute conventional logic gates by SOLGs, and set all output of the SO-ORs to logical 1. We now let the SOLC to  self-organize to satisfy the largest number of SO-ORs.

We have previously shown~\cite{DMM2} that SOLCs can be realized via standard (non-quantum) electronic components (we employ the realization described in 
Ref.~\cite{DMM2}, just slightly modified to deal with CNF formulas). 

One of the key components of SOLGs is the {\it dynamic correction module} we have designed to correct the inconsistent logic gate configurations. While the design and details of this component can be found in~\cite{DMM2}, we recall here its working
principle. The error correction module dynamically reads the voltages at the terminals of the gate, and injects a large current when the gate is in an inconsistent configuration, a small current otherwise.

The non-quantum electronic nature of SOLCs can be fully described by a system of {\it non-linear} ordinary differential equations of the type 
\begin{equation}
\dot{{\bf x}}(t) = {\bf F}({\bf x}(t)),\label{flow}
\end{equation}
where ${\bf x} = \{v_j, x_i\} \in X$ ($X$ is the phase space) is the collection of voltages, $v_j$, at the terminals and the internal state variables, $x_i$, of the electronic elements with memory; ${\bf F}$ is a system of nonlinear ordinary differential equations, representing 
the flow vector field~\cite{DMM2}.  	
	
We can then efficiently simulate them by 
numerical integration. Therefore, SOLCs are nothing other than dynamical systems. In this case, a solution of the problem we want to solve (e.g., the Max-SAT) employing a DMM is mapped into an equilibrium point of the dynamical system. The system is engineered in such a way that, starting from {\it any} initial condition (generally chosen at random) it evolves to converge into an equilibrium.

We have discussed in Ref.~\cite{DMM2} the relevant properties that the dynamical systems representing DMMs should have to behave in this way. Among them, an important feature, fundamental to guarantee the convergence, is that they are {\it point dissipative}~\cite{hale_2010_asymptotic}. This implies that the dynamical system has bounded orbits (no divergences), and it is endowed with an asymptotically stable global attractor, i.e., a compact set in the phase space that attracts any other point. This feature has also allowed us to prove that no chaotic behavior can emerge if equilibrium points are present~\cite{no-chaos}, as well as 
absence of periodic orbits~\cite{noperiod}.
Finally, the point dissipative property guarantees convergence to equilibrium {\it irrespective} of the initial conditions. 

We can finally summarize the power of these machines with the following hierarchical picture. DMMs use the {\it topology} of the internal connectivity of its 
elements to represent the problem to solve (this is called {\it information overhead} in Ref~\cite{DMM2}). Then, the collective state of the machine can manipulate all inputs, outputs and connecting variables in a massively-parallel fashion ({\it intrinsic parallelism}~\cite{DMM2}). 

In addition, the non-linearity of the dynamical 
system equations induces a transient instantonic phase with {\it long-range order}, both in space and time~\cite{topo}. This long-range order 
allows the system to converge exponentially fast to the equilibrium points that are associated to the approximations of optimization problems, by 
exploring a sub-space (that scales at most polynomially with input size) of the phase space. This sub-space is considerably smaller than the entire phase space itself~\cite{topo}. 

In fact, as briefly discussed in the main text, the particular realization of DMMs we have presented in this work (similar to the ones in Ref.~\cite{DMM2}) supports infinite-range correlations in 
the infinite input size limit, as shown in Ref~\cite{topo}. This enables an ideal scale-free behavior (namely one where the correlations {\it do not} decay) of the SOLC. This was derived analytically using topological field theory in Ref.~\cite{topo}, and can also be supported numerically from Fig. \ref{Fig_M_2} as follows. 

In order to simulate the system, we have employed a time-step size-controlled forward-integration scheme for the differential equations that describe it~\cite{Bulirsch2010}. Since the number of variables of the problem grows linearly with the input size because the number of gates grows only linearly, each time step to be simulated requires only a linear number of floating-point operations, and a memory linearly growing with input size. Then, the simulation time is just a linear function of the machine time. In Fig.~\ref{Fig_M_2} it is reported the same Fig.~\ref{Fig_3} but with the SOLC time (not normalized) on the x axis. It is evident that the relaxation of the system is independent of the input size (ideal scale-free scaling). This is a very interesting, and rare result for an extensive interconnected system. 
All these ingredients are necessary for the correct, efficient operation of a DMM.

The approximations to an optimization problem found by DMMs are very close to the global minimum of the problem, and this is guaranteed by the topology of the connectivity. 
This is clearly demonstrated in Fig.~\ref{Fig_M_1} where the unsatisfied clauses are plotted versus variables for  different simulation times, scaled linearly by the number of variables. While for the
random-Max-E3SAT both our solver and the CCLS approach the $0.4\%$ minimum, in the case of the hyper-Max-E3SAT, CCLS reaches a hard inapproximability limit of about $2\%$ for large instances. As expected, the delta-Max-E3SAT, instead, is a much worse case, and the inapproximability limit for CCLS is at about $3\%$.

In contrast, our non-combinatorial approach directly reaches the global minimum in all cases. Interestingly, our solver shows slightly better performances for the delta-Max-E3SAT (the most difficult of the three cases) as can be seen by taking a closer look at Fig.~\ref{Fig_M_1}.

\clearpage

\section{Supplemental information}

\section{A brief survey on MaxSAT Solvers}

As mentioned in the main text, there are two main (combinatorial) approaches to
solve or approximate the Max-SAT problem. The first is based on the exhaustive
exploration of the solution space and leads to the so-called ``complete''
solvers~\cite{gomes2008satisfiability,cocco2006}. The complete solvers use algorithms typically based on the
branch-and-bound approach~\cite{Optimization_book,heuristics_book} in which a greedy bound is first put on the optimum and then this is used to prune the resulting search tree.  Despite this pruning, they still scale exponentially with input size $|x|$ because they exhaustively search
a  space $\mathbb{Z}_{2}^{O(|x|)}=\{0,1\}^{O(|x|)}$. However, when the computation is finished, complete solvers are guaranteed to have found the global optimum of the Max-SAT.

Incomplete solvers \cite{gomes2008satisfiability,kautz2009incomplete}, in comparison, cannot guarantee the optimality of their solution as they do not explore the entire solution space. Instead they proceed by generating an initial assignment, and iteratively improving upon it. This trade-off allows them to find solutions, when they do, much more quickly than complete algorithms. In the most recent Max-SAT competition~\cite{MAXSAT_competition}, incomplete track solvers found solutions two orders of magnitude faster than complete track solvers in random and crafted benchmarks, and performed comparably on industrial benchmarks.

The quintessential incomplete solver is WalkSAT~\cite{kautz2009incomplete} which proceeds through a stochastic local search.  After an initial assignment is generated, an unsatisfied clause is selected and one variable from the clause has its assignment flipped. This will leave this clause satisfied but may alter the state of other clauses in which the variable occurs.  The procedure is continued for a specified number of steps or until a solution is found.  Most current local search solvers work similarly with various heuristics to select the next variable flip, utilize restarts and/or noise, and a host of other features.

We compared our solver, Falcon, with two of the best solvers from this years Max-SAT competition, CCLS~\cite{CCLS} and DeciLS~\cite{decils}. CCLS won the crafted track for unweighted Max-SAT and performs a local search (LS) with configuration checking (CC). Local search solvers will often retrace flips many times leading to an inefficient search.  Configuration checking keeps track of when neighboring variables have been flipped and only allows a variable to be flipped again when at least one of its neighbors has changed its assignment. DeciLS is an updated version of CnC-LS which won the industrial track for unweighted Max-SAT and combines a unit propagation based decimation (Deci) and local search (LS) with restarts.  An assignment is first generated through unit propagation-based decimation~\cite{cocco2006} in which conflicts are allowed, and the result is given to a local search for a specified number of steps. The process is then restarted and the best result of the previous search is used to guide the subsequent decimation and resolve conflicts.  This allows the solver to explore very different reasoning chains and areas of the solution space.

\section{Weighted partial Max-SAT}

In order to  more efficiently map a large number of maximization problems into
Max-SAT, it is sometimes useful to consider a variant:  {\it weighted} partial
Max-SAT~\cite{Optimization_book,heuristics_book}. Weighted partial Max-SAT is a version of Max-SAT for which 
a subset of clauses {\it must} be satisfied (``hard'' clauses), while the remaining clauses
(``soft'' clauses) may be weighted, and the sum of the weights of satisfied clauses must be maximized.
The Max-SAT is a particular case of the weighted partial MaxSAT in which all clauses are soft and have the same weight.

Because of the presence of hard clauses, the weighted partial Max-SAT is, in general, harder than the Max-SAT for all kind of solvers. In fact, this is one of the main reasons heuristics are often unable to find even approximations to those problems (see, e.g., Fig.~\ref{Fig_SI_3}--~\ref{Fig_SI_4}).  

Including weights and hard clauses in self-organizing logic circuits (SOLCs) is simple. Recalling that each OR gate representing a clause has attached at each terminal a dynamic correction module that injects a large current when the gate is in an inconsistent configuration, we can tune the maximum current allowed for each correction module in the following way. We set the maximum current injected by the dynamic correction modules connected to the SO-OR gates proportionally to the weights of the clauses. For the hard clauses we can set the maximum current injected by the dynamic correction modules connected to the hard SO-OR gates, larger than the sum of all maximum currents injected by the dynamic correction modules connected to all soft SO-OR gates connected to that hard SO-OR gate. This will guarantee that the hard clauses will have always the priority on the soft clauses. \\

\section{Comparison from the 2016 Max-SAT competition}

We have tested SOLCs on problems taken from the 2016 Max-SAT competition, and compared them against the results of the winners of each category of that competition. Even if the comparison is not completely fair because our code is written in MatLab while the other codes are written in compiled languages, and the benchmark is not the same because we ran on different processors (we ran all our simulations on an Intel Xeon E5-2680 v3 but used the same number of threads allowed in the Max-Sat competition) the results are still interesting.

\begin{figure}[t]
	\begin{center}
		\includegraphics[width=\columnwidth]{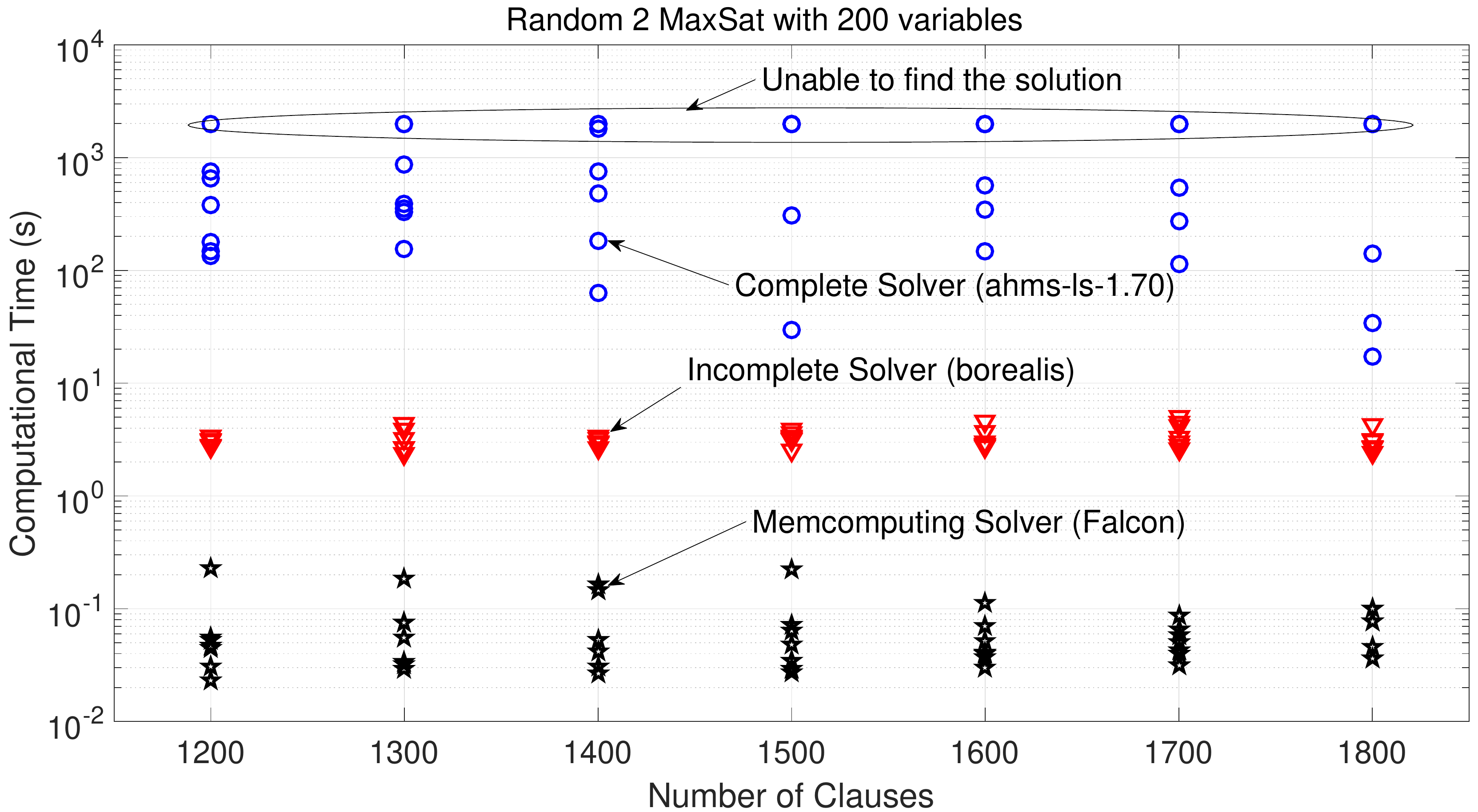}
	\end{center}
	\caption{Results from the 2016 Max-SAT competition for the random Max-2SAT problem compared with our memcomputing solver, Falcon.}%
	\label{Fig_SI_1}%
\end{figure}

\begin{figure}[t]
	\begin{center}
		\includegraphics[width=\columnwidth]{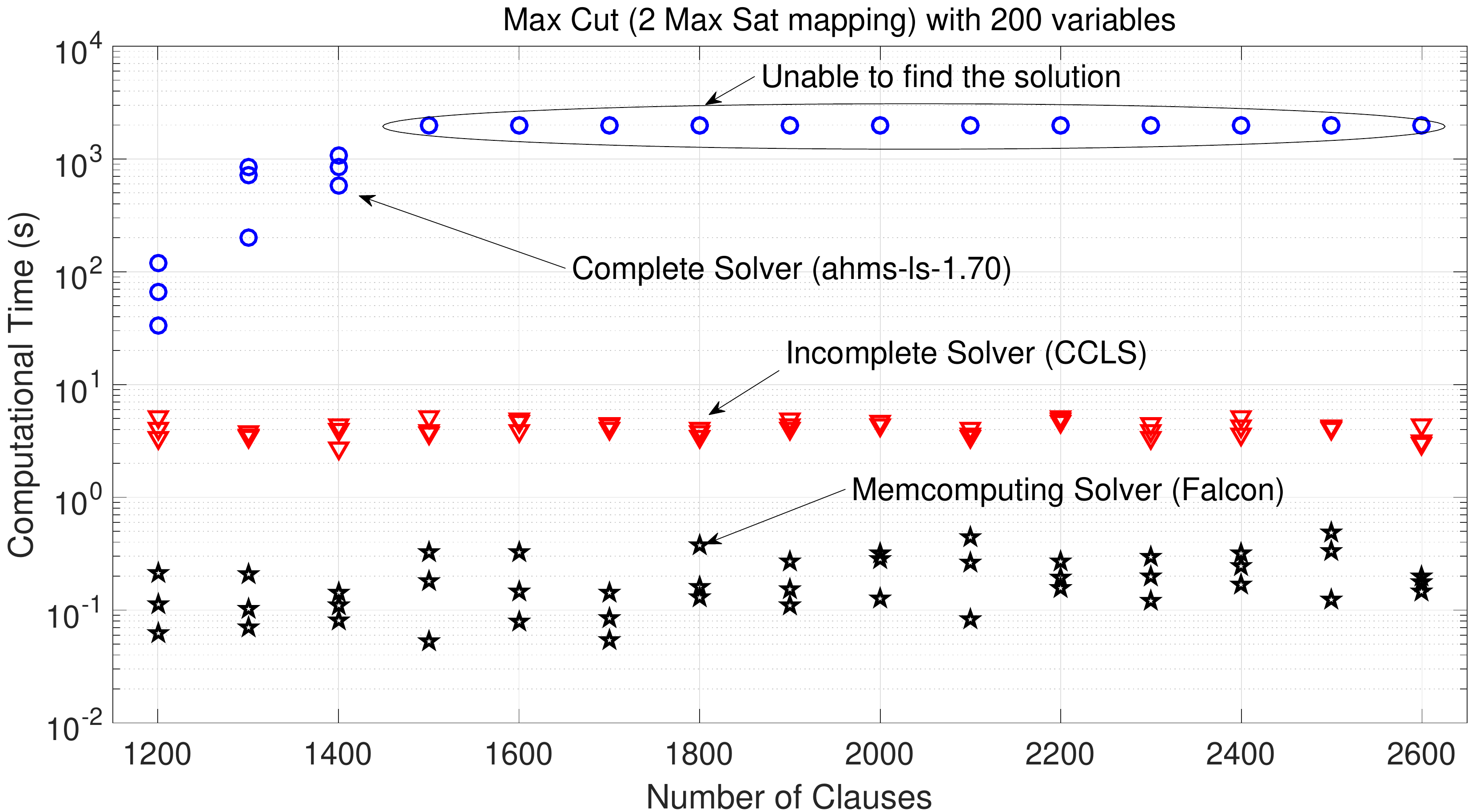}
	\end{center}
	\caption{Results from the 2016 Max-SAT competition for the Max-CUT problem compared with our memcomputing solver, Falcon.}%
	\label{Fig_SI_2}%
\end{figure}

In Figs.~\ref{Fig_SI_1} and~\ref{Fig_SI_2} we compare the random Max-2SAT and  random Max-CUT instances, which are non-weighted problems~\cite{Optimization_book}. In those cases the scaling is similar to the heuristics, but the absolute time is orders of magnitude lower.

\begin{figure}[t]
	\begin{center}
		\includegraphics[width=\columnwidth]{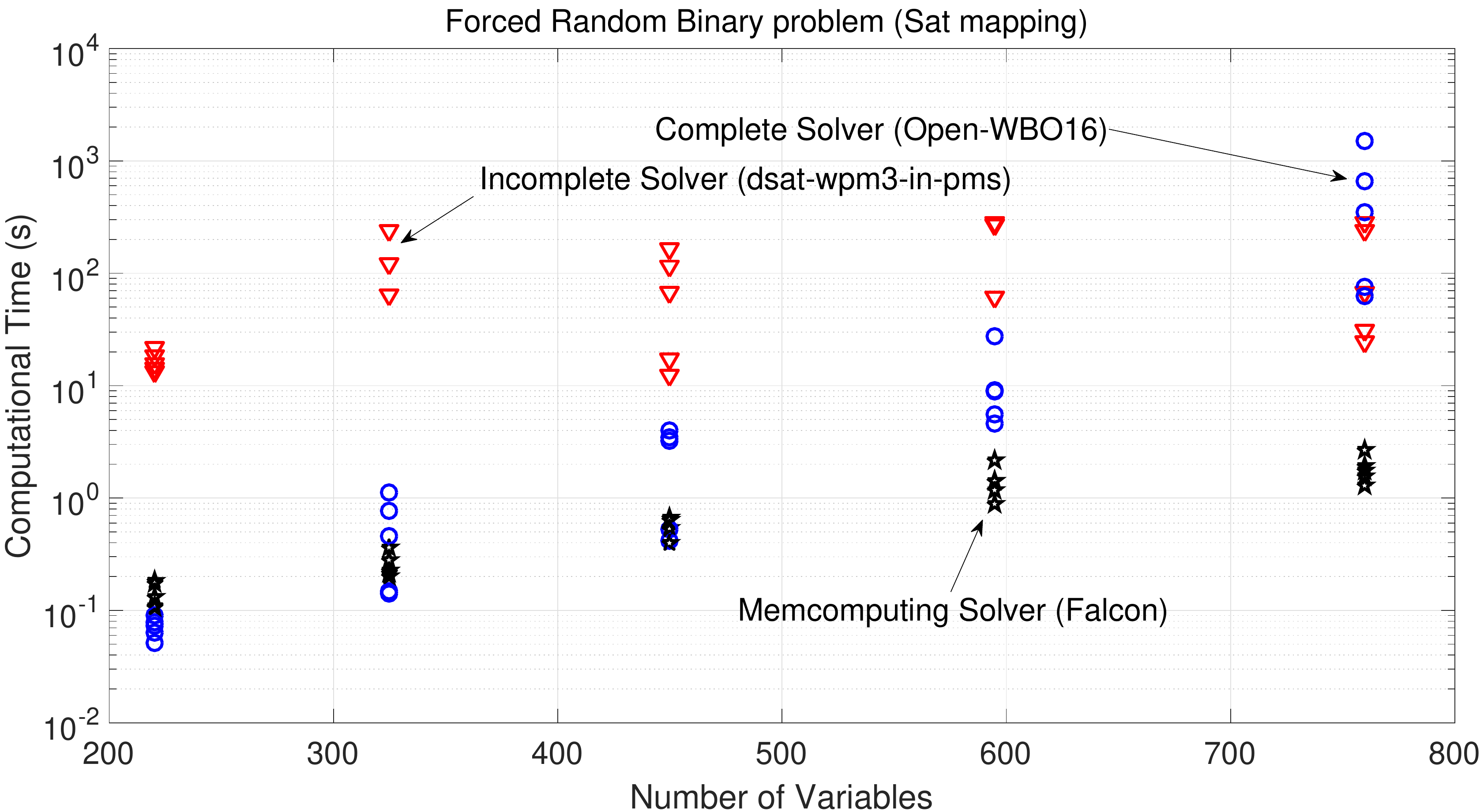}
	\end{center}
	\caption{Results from the 2016 Max-SAT competition for the Forced Random Binary problem compared with our memcomputing solver, Falcon.}%
	\label{Fig_SI_3}%
\end{figure}

\begin{figure}[t]
	\begin{center}
		\includegraphics[width=\columnwidth]{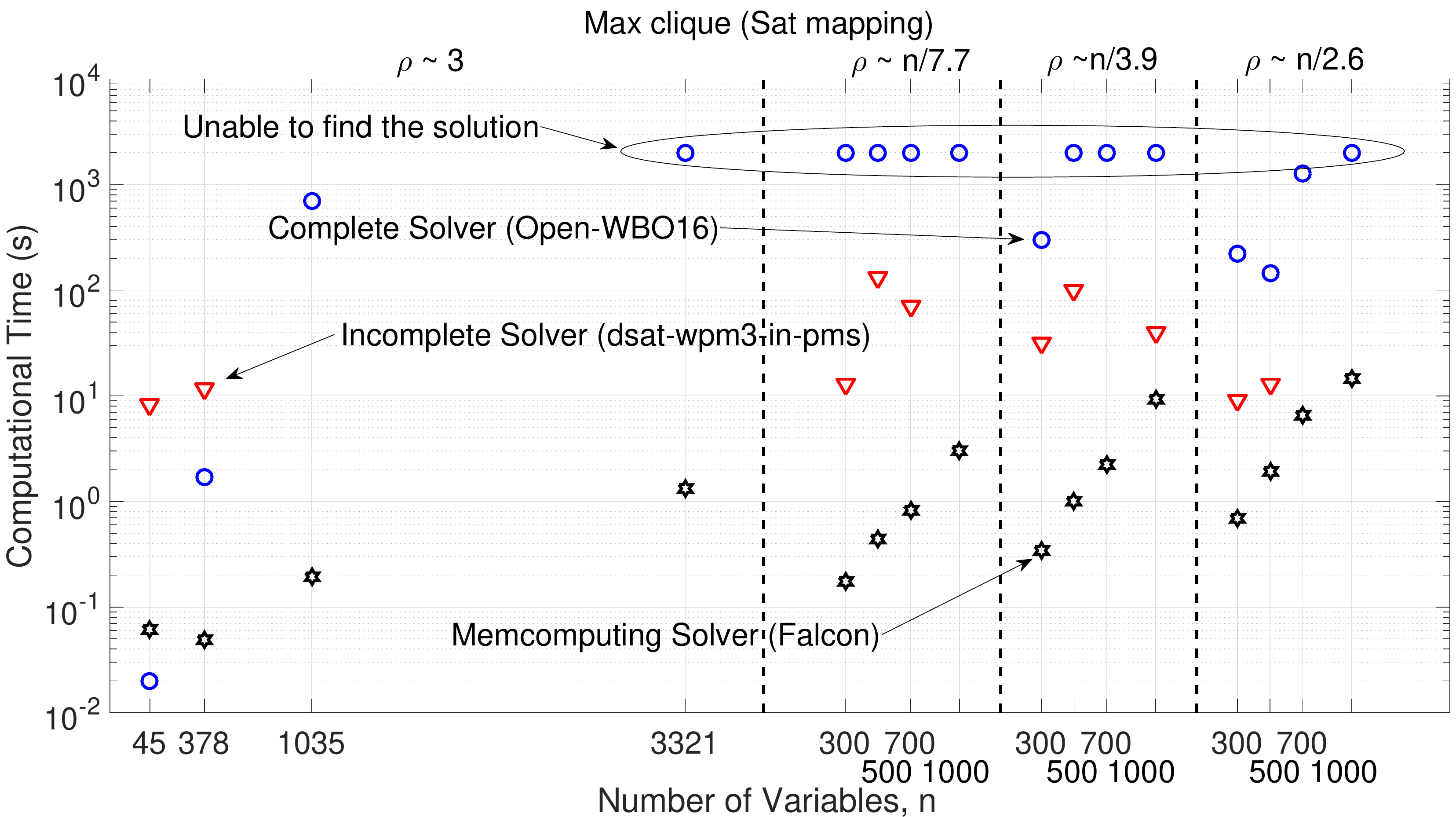}
	\end{center}
	\caption{Results from the 2016 Max-SAT competition for the Max-Clique problem compared with our memcomputing solver, Falcon.}%
	\label{Fig_SI_4}%
\end{figure}

Of more interest are the results of Figs.~\ref{Fig_SI_3} and~\ref{Fig_SI_4}. These correspond to two problems (called Forced Random Binary and Max Clique~\cite{Optimization_book}) that, when mapped, become weighted partial Max-SAT instances. As discussed, these are especially hard. In fact, 
oftentimes, the best heuristics cannot even find approximations because they were not able to satisfy all hard clauses, while our solver always does.

\bibliographystyle{naturemag}
\bibliography{SUSYref}

\begin{thebibliography}{10}
\expandafter\ifx\csname url\endcsname\relax
  \def\url#1{\texttt{#1}}\fi
\expandafter\ifx\csname urlprefix\endcsname\relax\def\urlprefix{URL }\fi
\providecommand{\bibinfo}[2]{#2}
\providecommand{\eprint}[2][]{\url{#2}}

\bibitem{DMM2}
\bibinfo{author}{Traversa, F.~L.} \& \bibinfo{author}{Di~Ventra, M.}
\newblock \bibinfo{title}{Polynomial-time solution of prime factorization and
  np-complete problems with digital memcomputing machines}.
\newblock \emph{\bibinfo{journal}{Chaos: An Interdisciplinary Journal of
  Nonlinear Science}} \textbf{\bibinfo{volume}{27}}, \bibinfo{pages}{023107}
  (\bibinfo{year}{2017}).

\bibitem{Optimization_book}
\bibinfo{author}{Christos H.~Papadimitriou, K.~S.}
\newblock \emph{\bibinfo{title}{Combinatorial Optimization}}
  (\bibinfo{publisher}{Dover Publications Inc.}, \bibinfo{year}{1998}).

\bibitem{Optimization_book_intro}
\bibinfo{author}{Edwin K. P.~Chong, S. H.~Z.}
\newblock \emph{\bibinfo{title}{An Introduction to Optimization}}
  (\bibinfo{publisher}{JOHN WILEY \& SONS INC}, \bibinfo{year}{2013}).

\bibitem{complexity_bible}
\bibinfo{author}{Garey, M.~R.} \& \bibinfo{author}{Johnson, D.~S.}
\newblock \emph{\bibinfo{title}{Computers and Intractability; A Guide to the
  Theory of NP-Completeness}} (\bibinfo{publisher}{W. H. Freeman \& Co.},
  \bibinfo{address}{New York, NY, USA}, \bibinfo{year}{1990}).

\bibitem{MAXSAT_competition}
\urlprefix\url{http://www.maxsat.udl.cat/16/index.html}.

\bibitem{gomes2008satisfiability}
\bibinfo{author}{Gomes, C.~P.}, \bibinfo{author}{Kautz, H.},
  \bibinfo{author}{Sabharwal, A.} \& \bibinfo{author}{Selman, B.}
\newblock \bibinfo{title}{Satisfiability solvers}.
\newblock In \bibinfo{editor}{Van~Harmelen, F.}, \bibinfo{editor}{Lifschitz,
  V.} \& \bibinfo{editor}{Porter, B.} (eds.) \emph{\bibinfo{booktitle}{Handbook
  of knowledge representation}}, vol.~\bibinfo{volume}{1},
  chap.~\bibinfo{chapter}{2}, \bibinfo{pages}{89--134}
  (\bibinfo{publisher}{Elsevier}, \bibinfo{year}{2008}).

\bibitem{kautz2009incomplete}
\bibinfo{author}{Kautz, H.~A.}, \bibinfo{author}{Sabharwal, A.} \&
  \bibinfo{author}{Selman, B.}
\newblock \bibinfo{title}{Incomplete algorithms}.
\newblock In \bibinfo{editor}{Biere, A.}, \bibinfo{editor}{Heule, M.} \&
  \bibinfo{editor}{van Maaren, H.} (eds.) \emph{\bibinfo{booktitle}{Handbook of
  satisfiability}}, vol. \bibinfo{volume}{185}, \bibinfo{pages}{185--203}
  (\bibinfo{publisher}{IOS press}, \bibinfo{year}{2009}).

\bibitem{heuristics_book}
\bibinfo{author}{Hromkovic, J.}
\newblock \emph{\bibinfo{title}{Algorithmics for Hard Problems: introduction to
  combinatorial optimization, randomization, approximation, and heuristics}}
  (\bibinfo{publisher}{Springer}, \bibinfo{year}{2010}).

\bibitem{Feige1998}
\bibinfo{author}{Feige, U.}
\newblock \bibinfo{title}{A threshold of ln n for approximating set cover}.
\newblock \emph{\bibinfo{journal}{Journal of the {ACM}}}
  \textbf{\bibinfo{volume}{45}}, \bibinfo{pages}{634--652}
  (\bibinfo{year}{1998}).

\bibitem{Hastad2001}
\bibinfo{author}{H{\aa}stad, J.}
\newblock \bibinfo{title}{Some optimal inapproximability results}.
\newblock \emph{\bibinfo{journal}{Journal of the {ACM}}}
  \textbf{\bibinfo{volume}{48}}, \bibinfo{pages}{798--859}
  (\bibinfo{year}{2001}).

\bibitem{UMM}
\bibinfo{author}{Traversa, F.~L.} \& \bibinfo{author}{Di~Ventra, M.}
\newblock \bibinfo{title}{Universal memcomputing machines}.
\newblock \emph{\bibinfo{journal}{IEEE Trans. Neural Netw. Learn. Syst.}}
  \textbf{\bibinfo{volume}{26}}, \bibinfo{pages}{2702} (\bibinfo{year}{2015}).

\bibitem{diventra13a}
\bibinfo{author}{Di~Ventra, M.} \& \bibinfo{author}{Pershin, Y.~V.}
\newblock \bibinfo{title}{The parallel approach}.
\newblock \emph{\bibinfo{journal}{Nature Physics}}
  \textbf{\bibinfo{volume}{9}}, \bibinfo{pages}{200} (\bibinfo{year}{2013}).

\bibitem{topo}
\bibinfo{author}{Di~Ventra, M.}, \bibinfo{author}{Traversa, F.~L.} \&
  \bibinfo{author}{Ovchinnikov, I.~V.}
\newblock \bibinfo{title}{Topological field theory and computing with
  instantons}.
\newblock \emph{\bibinfo{journal}{Ann. Phys. (Berlin)}}  (\bibinfo{year}{in
  press}).

\bibitem{13_amoeba}
\bibinfo{author}{Traversa, F.~L.}, \bibinfo{author}{Pershin, Y.~V.} \&
  \bibinfo{author}{Di~Ventra, M.}
\newblock \bibinfo{title}{Memory models of adaptive behavior}.
\newblock \emph{\bibinfo{journal}{IEEE Trans. Neural Netw. Learn. Syst.}}
  \textbf{\bibinfo{volume}{24}}, \bibinfo{pages}{1437--1448}
  (\bibinfo{year}{2013}).

\bibitem{DCRAM}
\bibinfo{author}{Traversa, F.~L.}, \bibinfo{author}{Bonani, F.},
  \bibinfo{author}{Pershin, Y.~V.} \& \bibinfo{author}{Di~Ventra, M.}
\newblock \bibinfo{title}{Dynamic computing random access memory}.
\newblock \emph{\bibinfo{journal}{Nanotechnology}}
  \textbf{\bibinfo{volume}{25}}, \bibinfo{pages}{285201}
  (\bibinfo{year}{2014}).

\bibitem{CCLS}
\bibinfo{author}{Luo, C.}, \bibinfo{author}{Cai, S.}, \bibinfo{author}{Wu, W.},
  \bibinfo{author}{Jie, Z.} \& \bibinfo{author}{Su, K.}
\newblock \bibinfo{title}{{CCLS}: An efficient local search algorithm for
  weighted maximum satisfiability}.
\newblock \emph{\bibinfo{journal}{{IEEE} Transactions on Computers}}
  \textbf{\bibinfo{volume}{64}}, \bibinfo{pages}{1830--1843}
  (\bibinfo{year}{2015}).

\bibitem{decils}
\bibinfo{author}{Cai, S.}, \bibinfo{author}{Luo, C.} \& \bibinfo{author}{Zhang,
  H.}
\newblock \bibinfo{title}{From decimation to local search and back: A new
  approach to maxsat}.
\newblock In \emph{\bibinfo{booktitle}{to appear in proceedings of
  International Joint Conference on Artificial Intelligence}}
  (\bibinfo{year}{2017}).

\bibitem{Mezard}
\bibinfo{author}{Mezard, M.} \& \bibinfo{author}{Montanari, A.}
\newblock \emph{\bibinfo{title}{Information, Physics, and Computation}}
  (\bibinfo{publisher}{Oxford University Press}, \bibinfo{year}{2009}).

\bibitem{mitchell1992hard}
\bibinfo{author}{Mitchell, D.}, \bibinfo{author}{Selman, B.} \&
  \bibinfo{author}{Levesque, H.}
\newblock \bibinfo{title}{Hard and easy distributions of sat problems}.
\newblock In \emph{\bibinfo{booktitle}{AAAI}}, vol.~\bibinfo{volume}{92},
  \bibinfo{pages}{459--465} (\bibinfo{year}{1992}).

\bibitem{kirkpatrick1994critical}
\bibinfo{author}{Kirkpatrick, S.}, \bibinfo{author}{Selman, B.} \emph{et~al.}
\newblock \bibinfo{title}{Critical behavior in the satisfiability of random
  boolean expressions}.
\newblock \emph{\bibinfo{journal}{Science-AAAS-Weekly Paper Edition-including
  Guide to Scientific Information}} \textbf{\bibinfo{volume}{264}},
  \bibinfo{pages}{1297--1300} (\bibinfo{year}{1994}).

\bibitem{cocco2006}
\bibinfo{author}{Cocco, S.}, \bibinfo{author}{Monasson, R.},
  \bibinfo{author}{Montanari, A.} \& \bibinfo{author}{Semerjian, G.}
\newblock \bibinfo{title}{Analyzing search algorithms with physical methods}
  (\bibinfo{year}{2006}).

\bibitem{barthel2002}
\bibinfo{author}{Barthel, W.} \emph{et~al.}
\newblock \bibinfo{title}{Hiding solutions in random satisfiability problems: A
  statistical mechanics approach}.
\newblock \emph{\bibinfo{journal}{Physical review letters}}
  \textbf{\bibinfo{volume}{88}}, \bibinfo{pages}{188701}
  (\bibinfo{year}{2002}).

\bibitem{ricci2001simplest}
\bibinfo{author}{Ricci-Tersenghi, F.}, \bibinfo{author}{Weigt, M.} \&
  \bibinfo{author}{Zecchina, R.}
\newblock \bibinfo{title}{Simplest random k-satisfiability problem}.
\newblock \emph{\bibinfo{journal}{Physical Review E}}
  \textbf{\bibinfo{volume}{63}}, \bibinfo{pages}{026702}
  (\bibinfo{year}{2001}).

\bibitem{jia2004}
\bibinfo{author}{Jia, H.}, \bibinfo{author}{Moore, C.} \&
  \bibinfo{author}{Selman, B.}
\newblock \bibinfo{title}{From spin glasses to hard satisfiable formulas}.
\newblock In \emph{\bibinfo{booktitle}{International Conference on Theory and
  Applications of Satisfiability Testing}}, \bibinfo{pages}{199--210}
  (\bibinfo{organization}{Springer}, \bibinfo{year}{2004}).

\bibitem{traversaNP}
\bibinfo{author}{Traversa, F.~L.}, \bibinfo{author}{Ramella, C.},
  \bibinfo{author}{Bonani, F.} \& \bibinfo{author}{Di~Ventra, M.}
\newblock \bibinfo{title}{Memcomputing {NP}-complete problems in polynomial
  time using polynomial resources and collective states}.
\newblock \emph{\bibinfo{journal}{Science Advances}}
  \textbf{\bibinfo{volume}{1}}, \bibinfo{pages}{e1500031}
  (\bibinfo{year}{2015}).

\bibitem{hale_2010_asymptotic}
\bibinfo{author}{Hale, J.}
\newblock \emph{\bibinfo{title}{Asymptotic Behavior of Dissipative Systems}},
  vol.~\bibinfo{volume}{25} of \emph{\bibinfo{series}{Mathematical Surveys and
  Monographs}} (\bibinfo{publisher}{American Mathematical Society},
  \bibinfo{address}{Providence, Rhode Island}, \bibinfo{year}{2010}),
  \bibinfo{edition}{2nd} edn.

\bibitem{no-chaos}
\bibinfo{author}{Di~Ventra, M.} \& \bibinfo{author}{Traversa, F.~L.}
\newblock \bibinfo{title}{Absence of chaos in digital memcomputing machines
  with solutions}.
\newblock \emph{\bibinfo{journal}{Phys. Lett. A}}
  \textbf{\bibinfo{volume}{381}}, \bibinfo{pages}{3255} (\bibinfo{year}{2017}).

\bibitem{noperiod}
\bibinfo{author}{{Di Ventra}, M.} \& \bibinfo{author}{Traversa, F.~L.}
\newblock \bibinfo{title}{Absence of periodic orbits in digital memcomputing
  machines with solutions}.
\newblock \emph{\bibinfo{journal}{Chaos: An Interdisciplinary Journal of
  Nonlinear Science}} \textbf{\bibinfo{volume}{27}}, \bibinfo{pages}{101101}
  (\bibinfo{year}{2017}).

\bibitem{Bulirsch2010}
\bibinfo{author}{R.~Bulirsch, J.~S.}
\newblock \emph{\bibinfo{title}{Introduction to Numerical Analysis}}
  (\bibinfo{publisher}{Springer}, \bibinfo{year}{2010}).

\end{thebibliography}
\end{document}